%% file: neurips_main.tex
\documentclass{article}

\usepackage[preprint]{neurips_2022}

\usepackage{natbib}
\setcitestyle{square,sort,comma,numbers}

\usepackage[utf8]{inputenc} 
\usepackage[T1]{fontenc}    
\usepackage{hyperref}       
\usepackage{url}            
\usepackage{booktabs}       
\usepackage{amsfonts}       
\usepackage{nicefrac}       
\usepackage{microtype}      
\usepackage{xcolor}         

\usepackage{times}
\usepackage{latexsym}
\usepackage{amsmath}
\usepackage{amsfonts}
\usepackage{multirow}
\usepackage{graphicx}
\usepackage{algorithm}
\usepackage{algpseudocode}

\title{Revisiting Attention Weights as Explanations from an Information Theoretic Perspective}

\author{%
    Bingyang Wen\\
  Infinity Lab \\
  Department of ECE \\
  Stevens Institute of Technology\\
  Hoboken, NJ 07030 \\
  \texttt{bwen4@stevens.edu} \\
  \And
  K.P. Subbalakshmi \\
    Infinity Lab \\
    Department of ECE \\
  Stevens Institute of Technology\\
  Hoboken, NJ 07030 \\
  \texttt{ksubbala@stevens.edu} \\
  \AND
    Fan Yang \\
    Department of ECE \\
  Stevens Institute of Technology\\
  Hoboken, NJ 07030 \\
  \texttt{fyang14@stevens.edu} 
}

\begin{document}

\maketitle

\input{content/abstract}

\input{content/introduction}

\input{content/preliminary}

\input{content/IB_attention}

\input{content/dataset}

\input{content/limitation}

\input{content/conclusion}

\bibliographystyle{plain}
\bibliography{anthology,explainability}


\appendix

\input{content/appendix}

\end{document}

%% file: content/abstract.tex
\begin{abstract}
Attention mechanisms have recently demonstrated impressive performance on a range of NLP tasks, and attention scores are often used as a proxy for model explainability. However, there is a debate on whether attention weights can, in fact, be used to identify the most important inputs to a model. We approach this question from an information theoretic perspective by measuring the mutual information between the model output and the hidden states. From extensive experiments, we draw the following conclusions: (i) \textit{Additive} and \textit{Deep} attention mechanisms are likely to be better at preserving  the information between the hidden states and the model output (compared to \textit{Scaled Dot-product}); (ii) ablation studies indicate that \textit{Additive} attention can actively learn to explain the importance of its input hidden representations; (iii) when attention values are nearly the same, the rank order of attention values is not consistent with the rank order of the mutual information
(iv) 
Using Gumbel-Softmax with a temperature lower than one, tends to produce a more skewed attention score distribution compared to softmax and hence is a better choice for explainable design; 
(v) some building blocks are better at preserving the correlation between the ordered list of mutual information and attention weights order (for eg. the combination of BiLSTM encoder and \textit{Additive} attention). Our findings indicate that attention mechanisms do have the potential to function as a shortcut to model explanations when they are carefully combined with other model elements.



\end{abstract}

%% file: content/introduction.tex
\section{Introduction}

Attention mechanisms have been used as a way to provide insight into the workings of deep learning models in several tasks~\cite{xuEtal15,choiEtal16,marEtal16,mulEtal18,thoEtal19}
However, there is a debate about whether attention weights are a good proxy for explanations. In~\cite{Jain&Wal19}, the authors argue that the attention weights cannot be used for explanations by (1) showing the inconsistency between the attention weights and other feature-importance measures; (2) showing an alternative attention distribution can yield similar results to those obtained by the original model. However, in  \cite{wie&pin19}, it was shown that the attention weights obtained from the original model are much more effective at prediction than the manually input decision-imitative attention weights. This result, while weakening the claim that ``attention is not explanation'', does not suggest that the attention mechanism is explanation. 

In this work, we continue to research this question, but from an information theoretic perspective. We focus on the encoder-attention-decoder structure~\cite{DzmEtal14}, which is one of the most common structures used in NLP tasks. 
Our analysis is different from those of \cite{Jain&Wal19,wie&pin19}, where the explainability of attention weights with respect to (w.r.t) the model inputs (e.g., token sequences) is studied. We focus on the attention mechanism itself and aim to evaluate if the attention mechanism can explain the importance of its inputs. Specifically, we look into representations of individual tokens, which are the input to the attention layer and the model output, and explore the ability of attention weights to serve as a proxy for the explanation. 

Our goal is to understand what role the attention mechanism plays in the encoder-attention-decoder framework from an information theoretic perspective. We ask the following questions: 
(1) how does the choice of the types of attention mechanism affect the amount of information that is encoded in the representations, which are inputs to the attention layer? 
(2) Can we identify specific combinations of "encoders"  and attention mechanisms that are best suited to build explainable models?
(3) Can attention mechanisms learn to assign higher attention values to representations that have higher mutual information (MI) w.r.t the model output? 

To answer these questions, we start by estimating the MI
between the model output and the inputs of the attention layer (e.g.,  representations of the tokens). The measured MI is representative of the amount of information about the outputs that is presented in each representation. By rank ordering the attention values and observing the variation of the corresponding MI values, we can estimate how closely the attention values track the information contained in the output values.

By measuring these relations across various types of encoder and attention mechanisms, we find that the attention scores obtained from \textit{Additive} attention~\cite{DzmEtal14} are the best proxy for explaining a model compared to \textit{Scaled Dot-Product}~\cite{AshEtal17} and \textit{Deep}~\cite{JohnEtal17} attention mechanisms. Additionally, we find that the combination of the BiLSTM encoder and \textit{Additive} attention~\cite{DzmEtal14} mechanism consistently yields attention score ranks that are positively correlated to the ranking of MI between the output and the representations. 

Then, based on the combination of BiLSTM and \textit{Additive} attention mechanism, we investigate the third question, namely, is the attention mechanism able to learn to assign higher attention scores to the representations with higher MI.
We find that the \textit{Additive} attention mechanism can indeed actively learn to assign higher scores to the input representations that contain more information about the output. Moreover, as in~\cite{wie&pin19}, we conduct an adversarial distribution study to assess the robustness of attention scores to such manipulations.
Our results show that, even under an adversarial manipulations  of the attention distribution (designed to produce similar output distribution while maximizing the Kullbach-Leibler (KL) distance from the 
the original distribution of the attentions), the attention weights 
still track the hidden representations with high MI.
These findings are encouraging that the attention mechanism has the potential to be used as a proxy for the model explanation. 
We summarize our contributions as follows:  

\begin{itemize}
    \item We analyze the workings of the attention mechanism from an information-theoretic perspective;
    \item We conduct extensive experiments across various types of encoder and attention mechanisms and find that the combination of BiLSTM and \textit{Additive} attention can be used to construct explainable models, since higher attention scores in this combination are always assigned to the representations that have the highest mutual information values with the model output. 
    \item We find that attention scores are unable to track informative inputs when their values are close. We also find that increasing the differences in attention values can improve their explainability in most cases.
    \item We find that the attention mechanisms can actively learn the important input representations;
    \item We conduct adversarial distribution analysis on attention weights and show that the attention modules are robust to such manipulations.
\end{itemize}





%% file: content/preliminary.tex
\section{Preliminary}
\label{sec:setting}
%

In this work, we focus on NLP tasks and follow a general encoder-attention-decoder structure.
%
We denote the model inputs $x \in \mathbb{R}^{T \times |\mathcal{V}|}$ as a sequence of tokens with length $T$, where each token is represented as a one-hot vector indicating a specific word in the vocabulary set $\mathcal{V}$. 
Firstly, the inputs are passed through the embedding layer to obtain a dense representation $x_{e} \in \mathbb{R}^{T \times d}$, where $d$ is the embedding size.
Then the embedded inputs are fed into the encoder (\textbf{Enc}) to produce $T$ hidden representations: $\mathbf{h} = \textbf{Enc}(x_e) \in \mathbb{R}^{T \times l}$, where $l$ is the size of the representation vector. 
Next, a set of $T$ attention scores $\hat{a} \in \mathbb{R}^{T}$ for the hidden representations $\mathbf{h}$ is calculated as: $\hat{a} = \textbf{Attn}(\mathbf{h}, Q)$ with some attention mechanisms $\textbf{Attn}$. 
Here, $Q \in \mathbb{R}^{l}$ is the query vector in some task\footnote{For example, this query vector can be either hidden representation of question in question and answering tasks or hypothesis in natural language inference} to obtain a query-related attention. Then, a context vector $c \in \mathbb{R}^{l}$ is calculated as the attention-weighted sum of all hidden representations: $c = \sum^{T}_{i=1} a_i h_i$, where $h_i$ is the $i^{th}$ hidden representation in $\mathbf{h}$ and $a_i \in \hat{a}$ is the attention score for $h_i$.
Finally, the decoder ($\mathbf{Dec}$) 
processes this context vector and produces the final output of the model $\hat{y} = \mathbf{Dec}(c) \in \mathbb{R}^{o}$, where $o$ denotes the size of the output and varies depending on the specific tasks for which the model is deployed (e.g. for binary classification $o=1$ and for multi-class classification $o > 1$ ). 

In the following sections, we will introduce the types of attention mechanisms (\textbf{Attn}), encoders (\textbf{Enc}), and decoders (\textbf{Dec}) that we consider. 

\subsection{Attention Mechanisms}
We consider three types of attention mechanisms for the $\mathbf{Attn}$ function described above:  \textit{Scaled Dot-Product}~\cite{AshEtal17} (denoted as \textit{Dot} in the rest of paper), \textit{Additive}~\cite{DzmEtal14}, and \textit{Deep}~\cite{JohnEtal17}. For the hidden representations $\mathbf{h}$, query vector $Q$ and softmax function $\sigma$, the \textit{Dot} attention scores are calculated as:
\begin{equation}
    \label{eqn:dot}
    \mathbf{Attn_{Dot}}(\mathbf{h}, Q) = \sigma(\frac{\mathbf{h}Q}{\sqrt{m}})
\end{equation}
The \textit{Additive} attention is defined as:
\begin{equation}
    \label{eqn:additive}
    \mathbf{Attn_{Add}}(\mathbf{h}, Q) = \sigma(W_1\text{tanh}(W_2\mathbf{h} + W_3Q)),
\end{equation}
where $W_1$, $W_2$ and $W_3$ are trainable weights. The \textit{Deep} attention is defined as:
\begin{equation}
    \label{eqn:deep}
    \mathbf{Attn_{Deep}}(\mathbf{h}, Q) = \sigma(D(\textit{W}_h\mathbf{h} + \textit{W}_QQ)) 
\end{equation}
where $D$ is a stack of $s$ fully connected layer with RELU activation: 
\begin{equation}
    \small
    \label{eqn:deep2}
    D(x)=W^{(s)}\text{RELU}(W^{(s-1)}\text{RELU}(...W^{(1)}\text{RELU}(x)))
\end{equation}

Note that the query vector is not considered in the original design of the \textit{Deep} attention mechanism. We extend it to process query vector by adding an extra weight $W_Q$. 

The above definitions consider the existence of query vectors. It is, however, not applicable in the classification tasks considered in our experiments, where there is no query vector. 
We hence adjust the usage of the above attention mechanisms and make them fit well in the case where only $h$ is given. For the decoder, we simply use a fully-connected layer with an activation function applied on its output (e.g., Sigmoid or Softmax) to map the context vector $c$ to the output $\hat{y}$, since our main target is to evaluate the explainability of attention modules and not performance.

For \textit{Additive} and \textit{Deep} attention, we simply remove the terms that contain query vectors: we remove $W_3Q$ from Eq.~\ref{eqn:additive} and for \textit{Dot}, we use the below functions instead:
\begin{equation}
    \label{eqn:dot_h_only}
    \mathbf{Attn_{Dot}}(\mathbf{h}) = \sigma(W^{D}_1\mathbf{h}),
\end{equation}
where $W^{D}_1$ is a trainable parameter.

\subsection{Encoder and Decoder}
In our experiments, we consider three types of encoder modules, which have different capacities for modeling contextual information. The encoders used in our experiments are Bi-LSTM, convolutional neural networks (CNN), and multi-layer perceptron (MLP). Bi-LSTM is believed to have the capacity to incorporate long-distance contextual information and can be useful for processing longer texts. 
CNN can only capture the information from a fixed number of neighboring elements where the number is decided by the kernel size of the convolutional layer. MLPs use an affine neural network to process each embedded input (i.e., one row in $x_e$) individually, and information from other time steps is never considered. 

For the decoder, we simply use a fully-connected layer with an activation function applied on its output (e.g., Sigmoid or Softmax) to map the context vector $c$ to the output $\hat{y}$, since our main target is to evaluate the explainability of attention modules and not performance.

%% file: content/IB_attention.tex
\section{Information Theoretic View of Attention}
%
%
%
In this section, we describe how we 
evaluate the explainability of the attention module from an information theory perspective. 
In Section~\ref{sec:information_bottlenect}, we describe the Information Bottleneck (IB) principle~\cite{tisEtal00} and another related method and discuss their relationship to the explainability of deep models in general. Inspired by these methods, we use mutual information to understand the explainability of the attention module, as described in Section~\ref{sec:our_method_mi}.


\subsection{Background}
\label{sec:information_bottlenect}
The information bottleneck (IB) method \cite{tisEtal00} was used to understand the workings of a general deep neural network (DNN)~\cite{tisEtal15}. The IB method showed that deep neural networks optimize each layer’s mutual information on input and output variables, resulting in a trade-off between compression and prediction and concludes that the optimal model transmits as much information as possible from the input $X$ to the output, $Y$, through a compressed representation (the hidden states in a DNN), $\hat{X}$, and notes that $X\rightarrow \hat{X} \rightarrow Y$ forms a Markov chain. 
%
They also showed that the stochastic gradient descent (SGD) algorithm followed the IB principle. Intuitively, the optimal $\hat{X}$ disregards all irrelevant parts in $X$ with respect to $Y$ and only keeps the relevant parts. 

Recently an approach called variational information bottleneck for interpretation (VIBI) was proposed for generating model agnostic explanations~\cite{banEtal21}, building on the IB theory. VIBI builds a trainable, post-hoc, model-agnostic explainer and consists of two parts: an explainer and a model approximator. The key difference between VIBI and IB is that VIBI considers instance-wise cognitive chunks (units that will act as explanations) and trades-off sufficiency with the brevity of the explanations.

Our work here is inspired by the above two approaches. In contrast to these approaches, however, we are neither trying to design an optimal DNN nor create a model agnostic explainer. We are interested in investigating the use of attention weights as a proxy for explanations since, if successful, attention weights can be a shortcut to explanations. We will do this by investigating the mutual information between the input and output of the attention modules under various settings.

\subsection{Analyzing the Attention Module using Mutual Information}
\label{sec:our_method_mi}
As mentioned earlier, previous works~\cite{Jain&Wal19,wie&pin19} assess the explainability of attention mechanisms by linking the attention scores to the importance of each token in the input sequence. 

\begin{figure}[ht]
    \centering
    \includegraphics[width=0.55\textwidth]{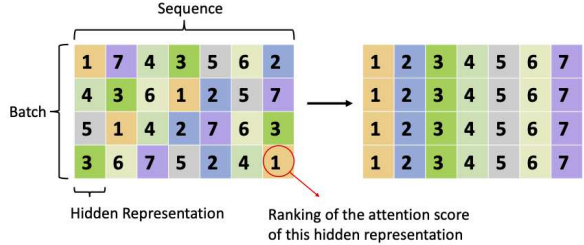}
    \caption{Diagram of collecting hidden representations for estimating mutual information. Each row represents a sequence of hidden representations of one data sample; Each cell represents one hidden representation vector where the number inside represents the rank of attention score assigned to this hidden state. On the right, it shows the collected hidden representations. }
    \label{fig:rankingMI}
\end{figure}

In this work, we evaluate if attention scores can be relied upon to explain the importance of the inputs to the attention mechanism. Specifically, we seek to understand how the attention scores are related to the information contained in the attention layer's input. Here, explainability is measured by the tie-up between attention scores and the mutual information between the input to the attention layer and the model output. 

The input to the attention layer is a sequence of hidden representations of the token sequence. Each hidden representation is associated with an attention score. We first rank these attention scores. We then create a vector of hidden representations formed by grouping the hidden representations for each data point with the same rank (see Fig~\ref{fig:rankingMI}). We then calculate the MI between the output and the vector of hidden representations for each rank.

%
%
%
This allows us to answer the question: ``does the attention mechanism assign the highest attention score to the most informative (w.r.t model output) input hidden representation?''

\subsubsection{Estimating Mutual Information}
To estimate MI, we need to consider hidden representations that are of the same attention rank as samples from a discrete distribution. 
%
Similar to the previous works~\cite{sajEtal18, voiEtal19}, we discretize the representations by clustering them into a large number of clusters. Then we use cluster labels instead of the continuous representations in the MI estimator. 







Specifically, we use $k$-means clustering to quantize the hidden representations and the output logits. We cluster the continuous representations into $N = 50-200$ clusters. The value of $N$ is selected as the minimal value that can ensure each centroid has at least two samples. For the binary classification task, we cluster the output logits into $N=5$ clusters. For multi-class classification tasks, such as NLI and Q\&A, we directly use the predicted categories. 

In practice, the sequences of hidden representations are of variable length. So, we pick the $k$ hidden representations with the highest attention scores. If $k$ is too small, then many hidden representations will not be included in the computation, and we may be estimating a local maximum of the MI; If $k$ is too large, some of the samples may not have sufficient length and hence will have to be removed from consideration when calculating the MI estimate. Hence the estimate could be noisy. Thus, we compromise on selecting $k$ as the minimum value that satisfies either of the following criteria: 10-th percentile of sequence length or the critical value of $k$ such that attention weights in 80\% of the selected hidden representations are larger than $10^{-5}$. The later criteria is based on the facts that extreme small attention weights will greatly reduce the gradients propagated to the corresponding hidden representation and hence is less likely to be ``informative''.  

Thus, we compromise on selecting $k$ so that the attention weights of 80\% of the $k^{th}$ hidden representations are larger than $10^{-5}$. The reason for this is that the hidden representations with extremely low attention weights are rarely updated to minimize the training error (due to small gradients) and hence have less likely to be informative.

%% file: content/dataset.tex
\section{Tasks and Datasets}
We follow the NLP tasks and datasets that are considered in ~\cite{Jain&Wal19} and their implementation for data pre-processing and model training. 
The NLP tasks include binary text classification, natural language inference and question answering. The statistics of all datasets are described in Table~\ref{tab:dataset}

\begin{table}[h]
\centering
\footnotesize
\scalebox{0.8}{
\begin{tabular}{ccc|cc|c}
\hline
\hline
Dataset & |V| & Avg. length & $\#$ Train & $\#$ Test & Performance (BiLSTM) \\
\hline
SST	       & 16175	& 19	& 3034 / 3321	& 863 / 862 	& 0.82\\
IMDB       & 13916  & 179 &  12500 / 12500 & 2184 / 2172  &  0.90  \\
ADR Tweets     & 8686 &  20  & 14446 / 1939 & 3636 / 487  & 0.86  \\
20 Newsgroups  & 8853 &  115  & 716 / 710 & 151 / 183  & 0.92 \\
AG News        & 14752 &  36  & 30000 / 30000 &  1900 / 1900 & 0.96  \\
Diabetes (MIMIC)    & 22316 &  1858  & 6381 / 1353 &  1295 / 319 & 0.89  \\
Anemia (MIMIC)      & 19743 &  2188  & 1847 / 3251 &  460 / 802 & 0.91 \\
\hline
bAbI (Task 1 / 2 / 3) & 40 &  8 / 67 / 421  & 10000 & 1000 & 0.98 / 0.74 / 0.57 \\
\hline
SNLI       & 20982 &  14  & 182764 / 183187 / 183416 & 3219 / 3237 / 3368  & 0.78  \\
\hline
\hline
\end{tabular}}
\caption{Dataset statistics. Generally, we follow the train-test split method used in ~\cite{Jain&Wal19}. For train and test size, the number of samples in each class is listed and separated by ``/''. The classes are seperated as 0/1 for binary classification (top), and 0 / 1 / 2 for NLI (bottom). Average length is in tokens. We report the accuracy of the BiLSTM encoder + \textit{Additive} attention mechanism. We note that results using CNN and  MLP encoders are comparable for classification though markedly worse for QA tasks.}
\label{tab:dataset}
\end{table}

\paragraph{Binary Text Classification} We use a total of 7 datasets for the text classification task. They include two sentiment analysis datasets from \textit{Standford Sentiment Treebank (SST)}~\cite{sst} and the \textit{IMDB Movie Reviews}~\cite{imdb}, where the target prediction is a binary variable to indicating positivity or negativity. The other five datasets include the \textit{Twitter Adverse Drug Reaction} dataset~\cite{tweet}, where the task is to identify if adverse drug reactions are mentioned in a tweet; \textit{20 Newsgroups} dataset\footnote{\url{https://archive.ics.uci.edu/ml/datasets/Twenty+Newsgroups}}, where the task is to distinguish if a news article is describing hockey or baseball; \textit{AG News Corpus}\footnote{\url{http://groups.di.unipi.it/~gulli/  AG\_corpus\_of\_news\_articles.html}} dataset, where the task is to discriminate between world and business articles;  Anemia and Diabetes datasets from \textit{MIMIC ICD9}~\cite{mimic}, where the tasks are to determine the type of Anemia (Chronic or Acute) and whether a patient is diabetic or not, respectively.

\paragraph{Natural Language Inference:} We consider the \textit{SNLI}~\cite{snli} dataset, where each sample contains a pair of sentences and a classification label to indicate the textual entailment within the sentence pair. There are three possible classification labels: entailment, contradiction, and neutral.

\paragraph{Question Answering} We use \textit{bAbI}~\cite{babi} dataset. Each sample in \textit{bAbI} contains a triplet of paragraph-question-answer, where the answer is one of the anonymized entities in the paragraph. The task requires the model to select an answer from the paragraph given the question. Specifically, we treat each of the three datasets presented in the original \textit{bAbI} datasets as an individual dataset. The differences between them are the number (from 1 to 3) of supporting statements (coherent with each other) for answering a question. 

\section{Results}
We now investigate if the attention scores are related to the amounts of information that the hidden states preserve about the model output using Kendall correlation between the ranks of mutual information and attention scores.
%
%

\begin{figure*}[ht!]
    \centering
    \includegraphics[scale=0.14]{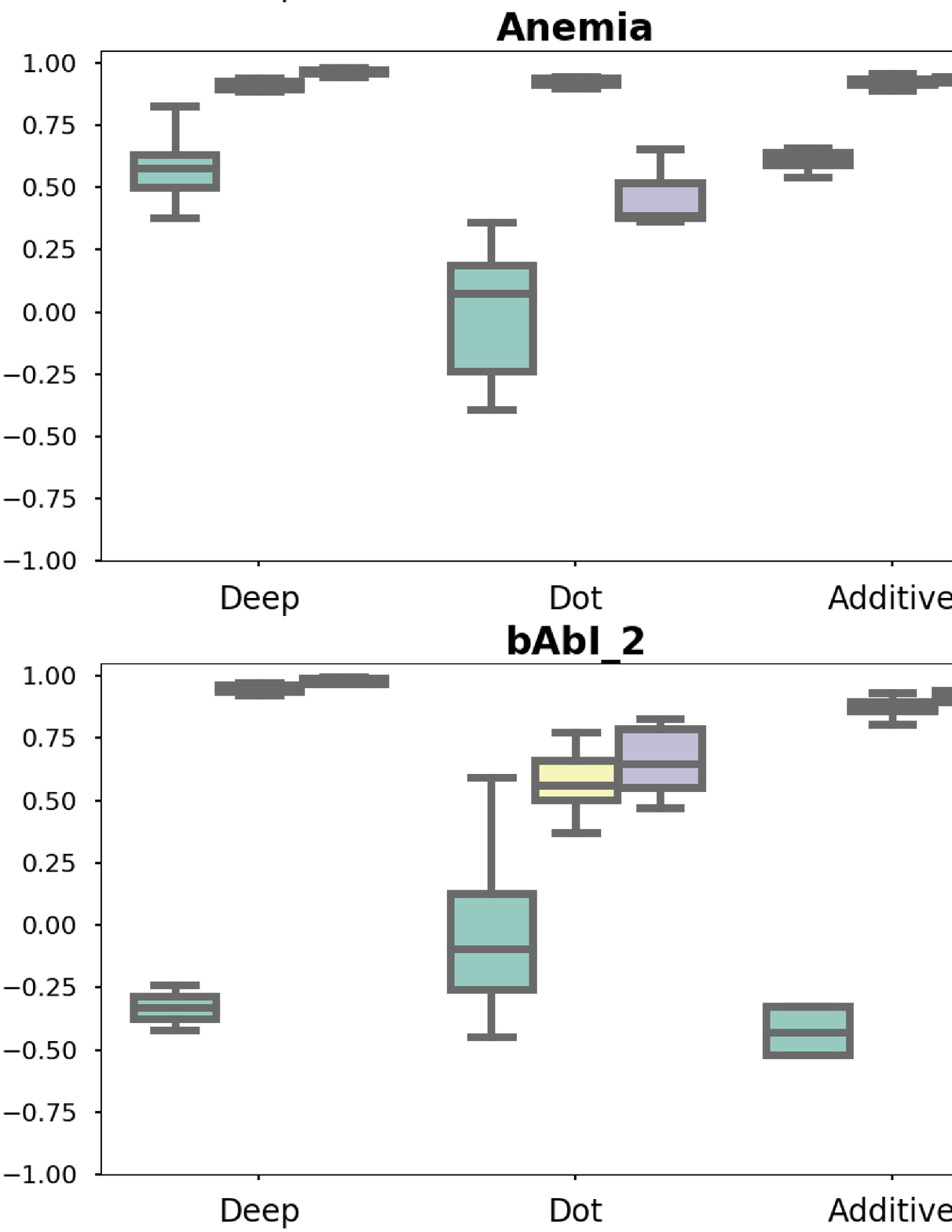}
    \caption{Box plots of \textbf{weighted Kendall correlation} between attention and mutual information for the representations (y-axis). X-axis denotes the types of the attention mechanism; Encoder variants are denoted by different colors. }
    \label{fig:kendall_mi_attn}
\end{figure*}

Kendall correlation~\cite{kendall38,shiEtal98} is a statistic
used to measure the ordinal association between two measured quantities. If the hidden representations with the highest attention scores also correspond to the highest value of mutual information between the input to the attention mechanism and the model output, we can conclude that there is a strong correlation between the attention scores and the mutual information. We measure Kendall correlation $\tau$ between $(m_1,...,m_k)$ and $(\bar{\hat{a}}_{1},...,\bar{\hat{a}}_{k})$, where $\bar{\hat{a}}_{i}$ is the mean of the $i^{th}$ largest attention scores across test dataset and $m_i$ is MI between the hidden representation corresponding to the $i^{th}$ value in the ordered attention score and the output. The highest value of $\tau$ is 1, and occurs when the rank orders of the two quantities are the same (i.e., most attended hidden states are the most informative about the output).

We use the weighted Kendall correlation~\cite{shiEtal98} with the attention scores (i.e., $\bar{\hat{a}}_{i}$) as the weights.
We do this to ensure that the mismatches at the high end (higher-ranked values) of the vector are penalized more than mismatches at the lower end of the vector.


Figure~\ref{fig:kendall_mi_attn} shows box plots of weighted Kendall correlation of 10 runs for each combination of encoder and attention mechanism\footnote{The details of the model settings and training are discussed in Appendix~\ref{sec:appendix_a}}. It illustrates that \textit{Deep} 
attention and \textit{Additive} attention achieves comparable results across all types of encoder on most classification datasets (except MLP encoder on Anemia dataset). For Q\&A and NLI tasks, we observe that $\tau$ values drop for all datasets, with the drops being more significant when the MLP encoder is used. The \textit{Dot} attention mechanism shows much lower 
correlations compared to \textit{Deep} and \textit{Additive} attention mechanisms on all datasets and all types of encoders. 

The average value of $\tau$ (across all datasets and all encoder types over 10 runs) for the \textit{Additive}, \textit{Deep} and \textit{Dot} attentions are $0.80$, $0.74$, and $0.45$, respectively. This indicates that among the three attention mechanisms, \textit{Additive} attention mechanism is the best at indicating the most informative hidden states. We also find that the combination of \textit{Additive} attention and BiLSTM has the best $\tau$ most of the time, with an average of  $0.92$. The $\tau$ values for CNN and \textit{Additive} attention, CNN and \textit{Deep} attention and BiLSTM and \textit{Deep} attention are also high at $0.88$, $0.83$ and $0.80$ respectively.

In the interest of reproducibility, we report the 
model accuracy of the BiLSTM and \textit{Additive} attention, averaged over $10$ different initialization seeds. For each random seed, we run the model 10 times. The results show negligible differences in model performance where the average absolute difference on Kendall $\tau$ is 0.02 and the maximum difference is 0.05 on \textit{AG News Corpus}. 

\subsection{Ablation}
In this section, we address the question: can the
attention module learn the highest mutual information representations?
%
We answer this question via an ablation study. 
In the \textbf{Fix Rep} setting, we study the effect 
of the attention mechanism by fixing the 
representations that are input to the attention module
and train the attention mechanism from scratch. 
In the \textbf{Fix Attn} setting, we fix the attention
scores, and train both the encoder and decoder from 
scratch to figure out if the BiLSTM would somehow 
``move'' the informative representations to the 
positions with high attention scores. In this section, we focus on the setting of BiLSTM + \textit{Additive} because this setting shows the most promising results so far. 

\begin{table}[t]

\centering
\caption{Differences in weighted Kendall correlation between normally trained BiLSTM+\textit{Additive} and the variants under special settings (\textbf{Fix Rep} and \textbf{Fix Attn}). $\Delta \tau = \tau_{\text{normal}} - \tau_{\text{special settings}}$.}
\scalebox{0.9}{
\begin{tabular}{ |ccc| } 
\hline
 \multirow{2}{*}{Dataset} & \multicolumn{2}{c|}{$\Delta \tau$}  \\ \cline{2-3}
 & Fixed Attn & Fixed Rep  \\ \hline
 SST         & 0.71 & -0.02 \\ \hline
 IMDB        & 0.16 & -0.01 \\ \hline
 AG News.    & 0.75 & -0.07 \\ \hline
 ADR Tweets. & 0.30 & 0.00 \\ \hline
 20News      & 1.32 & -0.07 \\ \hline
 Anemia      & 0.97 & -0.01 \\ \hline
 Diabetes    & 1.23 & -0.01 \\ \hline
 bAbI 1      & 1.70 & -0.02 \\ \hline
 bAbI 2      & 1.22 & -0.06 \\ \hline
 bAbI 3      & 0.66 & 0.00 \\ \hline
 SNLI        & 0.07  & -0.05 \\ \hline

\end{tabular}}

\label{tab:setting_ablation}
\end{table}

The changes in weighted Kendall correlation between the normally trained BiLSTM+\textit{Additive} and models trained in the above settings are reported in Table~\ref{tab:setting_ablation}. 
From this table, we notice that there is a large drop in the Kendall correlation in the \textbf{Fixed Attn} setting, whereas there is no drop in the Kendall correlations in the \textbf{Fix Rep} setting. Since the Kendall correlations were near perfect (i.e $\tau \sim 1$) for the normally trained BiLSTM+\textit{Additive}, this implies that the rankings 
of the MI and attention scores do not correlate well when the attention values are forced to a constant. This implies that 
the additive attention mechanism plays a vital role in tracking the representations with highest MI. Moreover, in the \textbf{Fix Rep}, we observe that the resultant weighted Kendall correlation either increase or remain the same. This indicates that BiLSTM does not add to the ability of keeping the correlation between MI and attention scores. This indicates that, under the same tasks, datasets and training objective functions, the additive attention modules learn to assign higher attention scores to the representations with higher MI w.r.t the output.
%

\subsection{Adversarial Model}
\label{sec:adv_model}
Both \cite{Jain&Wal19, wie&pin19} 
claim that the premise of ''attention is explanation" is that alternative attention weights ought to yield corresponding changes in output. They 
demonstrated that there exists an adversarial attention 
distribution that is very different from the
distribution of the attention weights in the normally trained
models that still yields similar output distribution.
In this section, we repeat our experiments in the 
Section~\ref{sec:our_method_mi} under adversarial 
settings to evaluate if the conclusions change. 

We consider the adversarial model proposed in~\cite{wie&pin19}
as the scenario considered in~\cite{Jain&Wal19} is less
realistic (a detailed discussion can be found in Section 2.1 
of \cite{wie&pin19}). Given the base model $M_b$, 
Authors of \cite{wie&pin19} train a model $M_a$ whose explicit goal is to provide similar prediction scores for each instance while distancing its attention distributions from those of $M_b$. 
The loss formula is defined as:
\begin{equation}
    \mathcal{L}(M_{a}, M_{b}) = \sum_{i=0}^{n}\text{TVD}(\hat{y}^{i}_{a} - \hat{y}^{i}_{b}) - \lambda \text{KL}(\hat{a}^{i}_{a} - \hat{a}^{i}_{b}),
\end{equation}
where $\hat{y}^{i}$ and $\hat{a}^{i}$ denote predictions and attention distributions for the $i^{th}$ sample, respectively and $\lambda$ is a hyperparameter. TVD stands for total variation distance, and KL stands for the Kullback–Leibler Divergence. In \cite{wie&pin19}, the above adversarial model is only tested on the classification datasets. 
To show how well the adversarial models are trained, we report the highest F1 scores of models, the divergence of attention distributions from the base, divergence of output distribution between the base and adversarial as well as their $\lambda$ setting, in Table~\ref{tab:adv_setting}.
In our work, we do the same. We also tried to extend their 
method of adversarial distribution generation for other datasets (Q\&A and NLI tasks), but it appears that this model does not work to produce different adversarial distributions for these tasks. Hence we test against adversarial distributions for classification tasks.

\begin{table}[h]
\centering
\caption{Evaluation and settings of the adversarial models. We use Jensen Shannon divergence (JSD) for measuring the divergence of output distributions.}

\scalebox{0.75}{
\begin{tabular}{ |ccccc| } 
\hline
 Dataset & $\lambda$ & Accuracy ($\uparrow$) & TVD ($\downarrow$) & JSD ($\uparrow$) \\ \hline
 SST         & 5.25e-4 & 0.82 & 0.036 & 0.316 \\ \hline
 IMDB        & 8e-4 & 0.90 & 0.021 & 0.482 \\ \hline
 AG News.    & 5e-4 & 0.96 & 0.009 & 0.664\\ \hline
 ADR Tweets. & 5e-4 & 0.87 & 0.001 & 0.194\\ \hline
 20News      & 5e-4 & 0.90 & 0.037 & 0.391\\ \hline
 Anemia      & 5e-4 & 0.91 & 0.020 & 0.384\\ \hline
 Diabetes    & 2e-4 & 0.89 & 0.021 & 0.464\\ \hline

\end{tabular}}

\label{tab:adv_setting}
\end{table}

The changes in weighted Kendall correlation between the 
normally trained BiLSTM+\textit{Additive} and models trained in the adversarial setting are reported in Table~\ref{tab:adv_diff}. We notice that there is not a big drop in the value of $\tau$. This implies that the attention values are still tied to the mutual information between the representation and the output. 

\begin{table}[h]
\centering
\caption{Differences in weighted Kendall correlation between normally trained BiLSTM+\textit{Additive} and its adversarially trained counterparts. $\Delta \tau = \tau_{\text{normal}} - \tau_{\text{adv}}$.}

\scalebox{1}{
\begin{tabular}{ |cc| } 
\hline
 Dataset & $\Delta \tau$  \\ \hline
 SST         & 0.04 \\ \hline
 IMDB        & 0.08 \\ \hline
 AG News.    & 0.05 \\ \hline
 ADR Tweets. & 0.02 \\ \hline
 20News      & 0.09 \\ \hline
 Anemia      & 0.02 \\ \hline
 Diabetes    & 0.05 \\ \hline

\end{tabular}}

\label{tab:adv_diff}
\end{table}


\subsection{Towards Designing Explainable Attention Networks}
We are interested to know when the attention weights lose their validity to indicate the most important hidden representations. We find that the attention weights for the models with lower Kendall $\tau$ tend to be more uniformly distributed than the models that achieve higher Kendall $\tau$. The finding shows that the attention weights are not accurate indicators of the quantity of information when attention weights are close. \textit{This finding would caution against using attention values as explanation without considering their relative magnitude.}


To further validate this hypothesis, we intervene in the attention scores generation process so that the differences between attention scores are larger. To do so, we simply replace the Softmax function $\sigma$ (in Eq.~\ref{eqn:dot}, \ref{eqn:additive}, \ref{eqn:deep}) in the attention layer with the Gumbel-Softmax function~\cite{JangEtal16} with a temperature of $0.8$. By setting the temperature to be smaller than one, the Gumbel-Softmax function is more likely to produce a “non-uniform” distribution than the Softmax function. Table~\ref{tab:mi_gumbel_diff} reports the statistics of changes in the Kendall $\tau$ for the models with the Gumbel-Softmax attention layers. We find that 85 out of 99 models can achieve higher or unchanged\footnote{We note that models with unchanged Kendall $\tau$ are originally near perfect ($\tau \sim 1$) Softmax attention layer. } Kendall $\tau$ when the Gumbel-Softmax is used to generate diverse attention values. 

These results answer our questions and suggest possible strategies to design explainable attention-based networks for NLP tasks. Moreover, we see that using Gumbel-Softmax along with \textit{Additive} attention mechanisms or on classification tasks improves the Kendall $\tau$ in most cases. However, such clear pattern is relatively weak in either \textit{Deep} and \textit{Dot} attentions or Q\&A and NLI tasks. 

\begin{table}[t]

\centering

\caption{Each entry reports the value of N/M, where M: total number of model settings. N: number of model settings (the combination of encoder and attention) for which the Kendall $\tau$ increases after using the Gumbel-Softmax attention layer.}

\scalebox{0.89}{
\begin{tabular}{ |c|ccc|c| } 
\hline
  Dataset & Deep & Dot & Additive & Average\\ \hline
  Classification & 20/21 & 19/21 & 20/21 & 59/63\\ 
  Q \& A & 6/9 & 6/9 & 8/9 & 20/27 \\ 
  NLI & 2/3 & 1/3 & 3/3 & 6/9 \\ \hline
  Average & 28/33 & 25/33 & 31/33 &  85/99 \\ \hline
  
\end{tabular}}

\label{tab:mi_gumbel_diff}
\end{table}

%% file: content/limitation.tex
\section{Limitation}
In order to complete the analysis of whether attention weights can indicate the most significant input to a model and hence provide an explanation of the model's performance in terms of its most significant inputs, we would need to tie the model input to the model output. So far, the analysis presented in this paper ties the output to the input to the last hidden layer. Further analysis will be necessary before we fully understand how to use attention weights as explanations.

%% file: content/conclusion.tex
\section{Conclusions and Future Work}
In this work we make a case that attention can in fact be 
used as a proxy for explanation of a model, when viewed from 
an information theoretic perspective for NLP classification tasks. We notice that (1) the \textit{Additive} and \textit{Deep} attention modules are the best at tracking the representations with highest mutual information between the 
model output and the input to the attention unit; (2) of all models tested, BiLSTM+\textit{Additive} attention seems to be the best at preserving the relationship between MI and attention weights; (3) when attention scores are more diverse, they are more representative of the importance of the inputs (4) additive attention mechanism is able to learn to assign highest attention weights to the representations with maximum MI and (5) Gumbel-Softmax is better at generating non-uniform distribution of attention values and hence better option for explainable model design for NLP classification tasks.

It would be interesting to explore more models in combination 
with the additive attention function. We believe this result 
help in constructing more explainable models using these 
building blocks.

%% file: content/appendix.tex
\section{Appendix A: Model Settings and Implementations}
\label{sec:appendix_a}
We acknowledge that most of our implementations of model training and datasets processing are originally derived from the work of~\cite{Jain&Wal19}, where the codes can be found at \url{https://github.com/successar/AttentionExplanation}. 
In our experiments, we consider three types of encoder and three types of attention mechanisms. Therefore, there are nine kinds of models obtained by combining the encoders and attentions. We use the same settings for the same type of encoders. 

For the BiLSTM encoder, we use one layer, bidirectional LSTM model with a hidden size of 128 for each direction. For the MLP encoder, we use a single hidden layer MLP with a hidden size of 128. For the CNN encoder, we set the sizes for the stacked filters as [1, 3, 5, 7, 15] and the number of output channels as 128. 

During training, we use Adam~\cite{kingma2014adam} optimizer with a learning rate of 0.001 and batch size of 32. We also set the weight decay coefficient to $10^{-5}$. The training process is stopped at the best performance on the validation dataset.